\documentclass[lettersize,journal]{IEEEtran}
\usepackage{amsmath,amsfonts}
\usepackage{algorithmic}
\usepackage{algorithm}
\usepackage{array}
\usepackage[caption=false,font=normalsize,labelfont=sf,textfont=sf]{subfig}
\usepackage{textcomp}
\usepackage{stfloats}
\usepackage{url}
\usepackage{verbatim}
\usepackage{graphicx}
\usepackage{cite}

\usepackage{authblk}
\usepackage{amssymb}
\usepackage{booktabs}
\usepackage{bbding}
\usepackage{hyperref}
\usepackage{ifpdf}
\usepackage{xspace}

\usepackage{multirow}
\usepackage{booktabs}
\usepackage{longtable}
\usepackage{xcolor}
\usepackage{textcomp}

\renewcommand\arraystretch{0.3}

\begin{document}

\title{Feature Activation Map: Visual Explanation of \\ Deep Learning Models for Image Classification}

\author{Yi Liao, Yongsheng Gao,~\IEEEmembership{Senior Member,~IEEE}, Weichuan Zhang,~\IEEEmembership{Member,~IEEE}
\thanks{This work is supported in part by the Australian Research Council under Industrial Transformation Research Hub Grant IH180100002 and Discovery Grant DP180100958 (Corresponding author: Yongsheng Gao). 

Yi Liao is with the School of Engineering and Built Environment, Griffith University, Brisbane, Queensland, 4111, Australia. (e-mail: yi.liao2@grifftihuni.edu.au). 

Yongsheng Gao is with the School of Engineering and Built Environment, Griffith University, Brisbane, Queensland, 4111, Australia. (e-mail: yongsheng.gao@griffith.edu.au).

Weichuan Zhang is with the Institute for Integrated and Intelligent Systems, Griffith University, Brisbane, Queensland, 4111, Australia. (e-mail: weichuan.zhang@griffith.edu.au).
}

}



\maketitle

\begin{abstract}
Decisions made by convolutional neural networks (CNN) can be understood and explained by visualizing discriminative regions on images. To this end, Class Activation Map (CAM) based methods were proposed as powerful interpretation tools, making the prediction of deep learning models more explainable, transparent, and trustworthy. However, all the CAM-based methods (e.g., CAM, Grad-CAM, and Relevance-CAM) can only be used for interpreting CNN models with fully-connected (FC) layers as a classifier. It is worth noting that many deep learning models classify images without FC layers, e.g., few-shot learning image classification, contrastive learning image classification, and image retrieval tasks. In this work, a post-hoc interpretation tool named feature activation map (FAM) is proposed, which can interpret deep learning models without FC layers as a classifier. In the proposed FAM algorithm, the channel-wise contribution weights are derived from the similarity scores between two image embeddings. The activation maps are linearly combined with the corresponding normalized contribution weights, forming the explanation map for visualization. The quantitative and qualitative experiments conducted on ten deep learning models for few-shot image classification, contrastive learning image classification and image retrieval tasks demonstrate the effectiveness of the proposed FAM algorithm.
\end{abstract}

\section{Introduction}
\IEEEPARstart{A}{lthough} deep learning models have achieved unprecedented success in a variety of computer vision tasks \cite{ref1AlexNet,ref2ResNet,ref3Dectection,ref4Segmentation,ref5DPSNEt}, the mystery hidden in the internal mechanisms of CNN has been attracting the attention from the computer vision community. Researchers attempt to unlock the reason why a network makes specific decisions and to visualize what regions on images are utilized for decision-making so that its decisions become explainable and trustworthy. To satisfy the demand, the class activation map (CAM) based methods \cite{ref6CAM, ref7GradCAM,ref8GradCAMplusplus,ref9LayerCAM,ref10AblationCAM,ref11XCAM,ref12ScoreCAM,ref13RelevanceCAM,ref14LFICAM,ref15LIFTCAM,ref16EigenCAM} has been proposed. They are viewed as powerful interpretation tools to enable the decisions of deep neural networks to be more transparent. CAM-based methods aim to generate visual explanation maps by linearly combining feature importance coefficients with activation maps (the matrix of each channel in a feature map). Although different methods obtain feature importance coefficients in different ways, they all inevitably resort to the fully connected layer (FC layer) as a classifier \cite{ref6CAM, ref7GradCAM,ref8GradCAMplusplus,ref9LayerCAM,ref10AblationCAM,ref11XCAM,ref12ScoreCAM,ref13RelevanceCAM,ref14LFICAM,ref15LIFTCAM,ref16EigenCAM}. The reason why the CAM-based approaches can correctly and effectively visualize the discriminative regions of the target objects is that they utilize the feature information of target category. The feature information is acquired through training data and stored in FC layers as a classifier \cite{ref16EigenCAM}. A FC layer, which plays the role of classification in deep neural network architecture, is also called as a linear classifier. Although various backbone networks (e.g., ResNet\cite{ref2ResNet}, VGG \cite{ref17VGG}, InceptionNet \cite{ref18InceptionNet}, and Vision Transformer \cite{ref19ViT}) perform as feature extractors, FC layers are commonly employed to identify a testing sample. It is worth noting that there are two constraints when FC layers are chosen as classifiers. First, FC layers need to be trained well before testing. Second, the training data and testing samples must have the same class domain.

Many deep learning networks in performing computer vision tasks do not use FC layers. Classifying testing samples via similarity comparison is a classification paradigm that does not rely on FC layers \cite{ref20NPID,ref21MatchingNet,ref22PNet}. Because the paradigm gets rid of the two aforementioned restrictions, it can be broadly adopted for more diverse recognition tasks than FC layer-based paradigm, from unsupervised learning tasks \cite{ref7GradCAM} to train-test domain shift tasks (e.g., few-shot learning \cite{ref21MatchingNet,ref22PNet}, and image retrieval \cite{ref24MSLoss} tasks). However, the existing CAM-based methods rely on FC layers as classifiers. They fail to function in theory and cannot be used for visualizing salient regions of deep learning models that use FC-free paradigm. For example, due to the constraint caused by FC layers, all CAM-based methods can only visualize salient regions of images from seen classes, they cannot be applied to visualizing explanation maps for unseen classes that are disjoint from training data in train-test domain shift tasks.

In this work, we fill this critical gap by presenting a novel model-agnostic visual explanation algorithm named feature activation map (FAM), for visualizing which regions on images are utilized for decision-making by those deep learning networks without FC layers. The main contributions of this paper can be summarized as follows:

1.~A novel visual explanation algorithm, feature activation map, is proposed that can, for the first time, visualize FC-free deep learning models in image classification.

2.~FAM can provide saliency maps for any deep learning networks that use similarity comparison classification paradigm without the constraint to the model architecture.

3.~The experiments demonstrate the effectiveness of the proposed method in explaining what regions are utilised for decision-making by various deep learning models for few-shot learning, contrastive learning and image retrieval tasks.
\section{Related Work}
\subsection{Similarity Comparison-based Classification}
Because extending the training set to cover endless classes in the world is infeasible, the main goal of unsupervised learning is to learn features that are transferrable \cite{ref23MoCo}. However, FC layers as a classifier are not generalized to new classes \cite{ref20NPID}. Wu et al. \cite{ref20NPID} propose a memory bank to replace FC layers, which construct a non-parametric softmax classifier by using cosine similarity between two feature embeddings. The neighbours with the top k largest similarities are used to make the prediction via voting. A memory bank is widely employed for downstream classification tasks in contrastive learning \cite{ref20NPID,ref23MoCo}, even in supervised learning classification tasks \cite{ref25PointBert}. Few-shot image classification \cite{ref21MatchingNet,ref22PNet} aims to learn transferable feature representations from the abundant labeled training data in base classes, such that the feature can be easily adapted to identify the unseen classes with limited labelled samples. It is worth noting that the unseen class is disjoint from the based classes, so the FC layers fail to play the role of classifier. To this end, classification is implemented by comparing the similarity between the support and query images. The cosine similarity \cite{ref26CAN} and Euclidean distance \cite{ref27FRN} are two popular similarity metrics in few-shot image classification. Pair-based deep metric learning methods are built on pairs of samples, aiming to minimize pairwise cosine similarities in the embedding space, they have been applied to the field of image retrieval \cite{ref24MSLoss,ref28MarginLoss}. When image retrieval deep-learning models make decisions via the best similarity, the retrieval tasks can be viewed as classification tasks.
\subsection{CAM-based Methods}
CAM-based methods can be briefly categorized into CAM, Gradient-based methods, and Gradient-free methods. Zhou et al. \cite{ref6CAM} proposed the original CAM which produces class-discriminative visualization maps by linearly combining activation maps at the penultimate layer with the importance coefficients that are the FC weights corresponding to the target class. CAM is constrained to the model architecture where the model must consist of global average pooling (GAP) layer and one FC layer as its classifier. Gradient-based methods include Grad-CAM \cite{ref7GradCAM}, Grad-CAM++ \cite{ref8GradCAMplusplus}, Layer-CAM \cite{ref9LayerCAM}, and XGrad-CAM \cite{ref11XCAM}. Grad-CAM \cite{ref7GradCAM} was proposed to explain CNNs without the limit to GAP layer as required by CAM. The importance coefficients are computed by averaging all first-order partial derivatives of the class score with respect to each neuron at activation maps. Although Grad-CAM++ \cite{ref8GradCAMplusplus}, Layer-CAM \cite{ref9LayerCAM}, and XGrad-CAM \cite{ref11XCAM} adopt various strategies to generate the importance coefficients for visual explanation maps, the calculation of the coefficients always uses the first order partial derivatives of the class score, which is not absolutely computed without the assistance of FC layers as a classifier. The reason that gradient-based methods only work for deep models with FC layers as classifiers is that decision boundary is learned by using FC layer \cite{ref16EigenCAM} to ensure the gradient-based methods can correctly highlight the discriminative class feature. However, as a network deepens, gradients become noisy and tend to diminish due to the gradient saturation problem \cite{ref13RelevanceCAM}, using unmodified raw gradients results in failure of localization for relevant regions \cite{ref12ScoreCAM,ref13RelevanceCAM,ref15LIFTCAM}. To avoid the gradient issues, gradient-free methods \cite{ref10AblationCAM, ref12ScoreCAM,ref13RelevanceCAM,ref14LFICAM,ref15LIFTCAM} are proposed, but gradient-free methods still rely on FC layers as a classifier for explainable visualization. The detailed analysis will be formally stated in Section \ref{sub3problem}. 
\section{Statement of Problem}
\label{sub3problem}
In this section, we theoretically state that FC layers as a classifier are indispensable for CAM-based methods to generate the explanation maps. 
\subsection{Class Activation Map}
CAM\cite{ref6CAM} requires that CNN architectures must include the global average pooling (GAP) layer and the FC layer as a classifier. Let A be the feature map as the output of the final convolutional layer, which consists of a series of activation maps from $A^1$ to $A^N$. Thus, CAM is defined as
\begin{equation}\begin{aligned}
\label{eq1}
L_{CAM}^{c}=\sum_{n=1}^{N}w_{n}^{c}A^{n},
\end{aligned}\end{equation}
where $A^n$ is the $n$-th activation map of $A$, $N$ denotes the number of channels of $A$, and $w_n^c$ is the weight corresponding to the class $c$ from the FC layer as a classifier. According to the definition of CAM, FC weights corresponding to class $c$ are directly used as the importance coefficients, CAM cannot be obtained without FC layer.
\subsection{Gradient-based Methods}
According to \cite{ref7GradCAM}, Grad-CAM is defined by the following,
\begin{equation}\begin{aligned}
\label{eq2}
L_{Grad-CAM}^{c}=\text{ReLU}\bigg(\sum_{n=1}^{N}a_{n}A^{n}\bigg),
\end{aligned}\end{equation}
and the importance coefficients are formulated as,
\begin{equation}\begin{aligned}
\label{eq3}
a_{n}=\frac{1}{D}\sum_{i}\sum_{j}\frac{\partial{y^c}}{\partial{A_{i,j}^{n}}},
\end{aligned}\end{equation}
where $D$ is the number of all units on $A^{n}$,~$y^{c}$ is the classification score for class $c$, and $A_{i,j}^{n}$ refers to the activation value at location ($i$,$j$) on $A^{n}$. It is worth noting that Grad-CAM++~\cite{ref8GradCAMplusplus}, Layer-CAM~\cite{ref9LayerCAM}, and XGrad-CAM~\cite{ref11XCAM} all contain the same partial derivative $\frac{\partial{y^c}}{\partial{A_{i,j}^{n}}}$ as Grad-CAM. As an input image $X$ is embedded into a feature vector $Z=[z_1,z_2,...,z_N]^{T}$, the class $c$'s score $y^c$ is obtained by
\begin{equation}\begin{aligned}
\label{eq4}
y^c=\sum_{n\text{=}1}^{N}z_{n}w_{n}^{c}+\text{bias}.
\end{aligned}\end{equation}
Following Chain Rule, we can have the following equation,
\begin{equation}\begin{aligned}
\label{eq5}
\frac{\partial{y^c}}{\partial{A_{i,j}^{n}}}=\sum_{n\text{=}1}^{N}\bigg(w_{n}^{c}\frac{\partial{z_{n}}}{\partial{A_{i,j}^{n}}}\bigg),
\end{aligned}\end{equation}
where $W^c=[w_1^c,w_2^c,...,w_N^c]$ is the weight vector of FC-layers corresponding to the class $c$. Eq.~(\ref{eq5}) shows that Gradient-based methods definitely rely on FC layers as a classifier. 
\subsection{Gradient-free Methods}
Ablation-CAM \cite{ref10AblationCAM} and Score-CAM \cite{ref12ScoreCAM} are respectively defined by the following, 
\begin{equation}
\begin{aligned}
\label{eq6AblationCAM}
L_{Ablation-CAM}^{c}=\text{ReLU}\bigg(\sum_{n=1}^{N}\frac{y^c-y_{n}^{c}}{y^{c}}A^{n}\bigg),
\end{aligned}
\end{equation}
where $y_n^c$ is the score with the absence of $A^n$, and
\begin{equation}
\begin{aligned}
\label{eq7ScoreCAM}
L_{Score-CAM}^{c}=\text{ReLU}\bigg(\sum_{n=1}^{N}(\text{softmax}(F^{c}(X')-F^{c}(X)))A^{n}\bigg),
\end{aligned}
\end{equation}
where $X'=X\circ\text{norm}(\text{up}(A^n))$ and $\circ$ denotes Hadamard Product, $F^c$ and $X$ denote a CNN model and an input image respectively, so $y^c=F^c (X)$. The function $\text{up}(\cdot)$ denotes the up-sampling operation that scales $A^n$ to the size of the image $X$, $\text{norm}(\cdot)$ and $\text{softmax}(\cdot)$ indicate the max-min normalization function and softmax function respectively. It is worth noting that Eqs.~(\ref{eq6AblationCAM}) and (\ref{eq7ScoreCAM}) include the same component $y^c$. Eq.~(\ref{eq4}) shows $y^c$ cannot be computed without the weight $w_n^c$. Relevance-CAM \cite{ref13RelevanceCAM} uses the index of $y^c$ to define the relevance score of the final layer by
\begin{equation}
\begin{aligned}
\label{relevance_score}
R_{\widetilde{n}}^{(L)}=
\begin{cases}
v_t^{(L)}&{\widetilde{n}}=t\\
-\frac{v_t^{(L)}}{\widetilde{N}-1}&\text{otherwise},
\end{cases}
\end{aligned}
\end{equation}
where $v_t^{(L)}$ denotes the model output value for target class index $t$ on $L$-th layer and $\widetilde{N}$ denotes the number of classes. $\widetilde{N}$ is derived from the shape of FC layers as a classifier. For train-test domain shift tasks, $\widetilde{N}$ is not available because test classes are unseen. To obtain the importance coefficient of the $n$-th channel activation map, the relevance score of the final layer should be backward propagated to the intermediate convolutional layer through the FC layer as a classier. The propagation rule is written as the following equation,
\begin{equation}
\begin{aligned}
\label{z_rule}
R_i=\sum_{j}\frac{{A^i}{w_{ij}^{+}}}{\sum_{i}{{A^i}{w_{ij}^{+}}}}R_j,
\end{aligned}
\end{equation}
where $R_i$, $R_j$ denote the $i$-th layer relevance score and the $j$-th layer relevance score, respectively, $A^i$ denotes the activation output of the $i$-th layer, and $w_{ij}^{+}$ denotes the positive part of the weight between the $i$-th and the $j$-th layers. Because backward propagation always passes firstly through the FC layers as a classifier, $w_{ij}^{+}$ should always use the weight of the FC layers as a classifier. LIFT-CAM \cite{ref15LIFTCAM} can approximate the SHAP values \cite{ref29SHAPValue} as the importance coefficients during backward propagation by using $\alpha_{n}^{\text{lift}}$, which is calculated by 
\begin{equation}
\begin{aligned}
\label{lift_cam}
\alpha_{n}^{\text{lift}}=\sum_{(i,j)\in\land}{A^n_{i,j}w^c_{(n,i,j)}},
\end{aligned}
\end{equation}
where $w^c_{(n,i,j)}$ indicates the network weight corresponding to $A^n_{i,j}$ for class $c$~and~$\land$ denotes the size of $A^n$. Eq.~(\ref{lift_cam}) requires that the network architecture should have FC layers as a classifier to provide the weight corresponding to each target class\cite{ref15LIFTCAM}.
\begin{figure*}
\centering
\setlength{\abovecaptionskip}{5pt}
\setlength{\belowcaptionskip}{5pt}
\includegraphics[width=2.1\columnwidth]{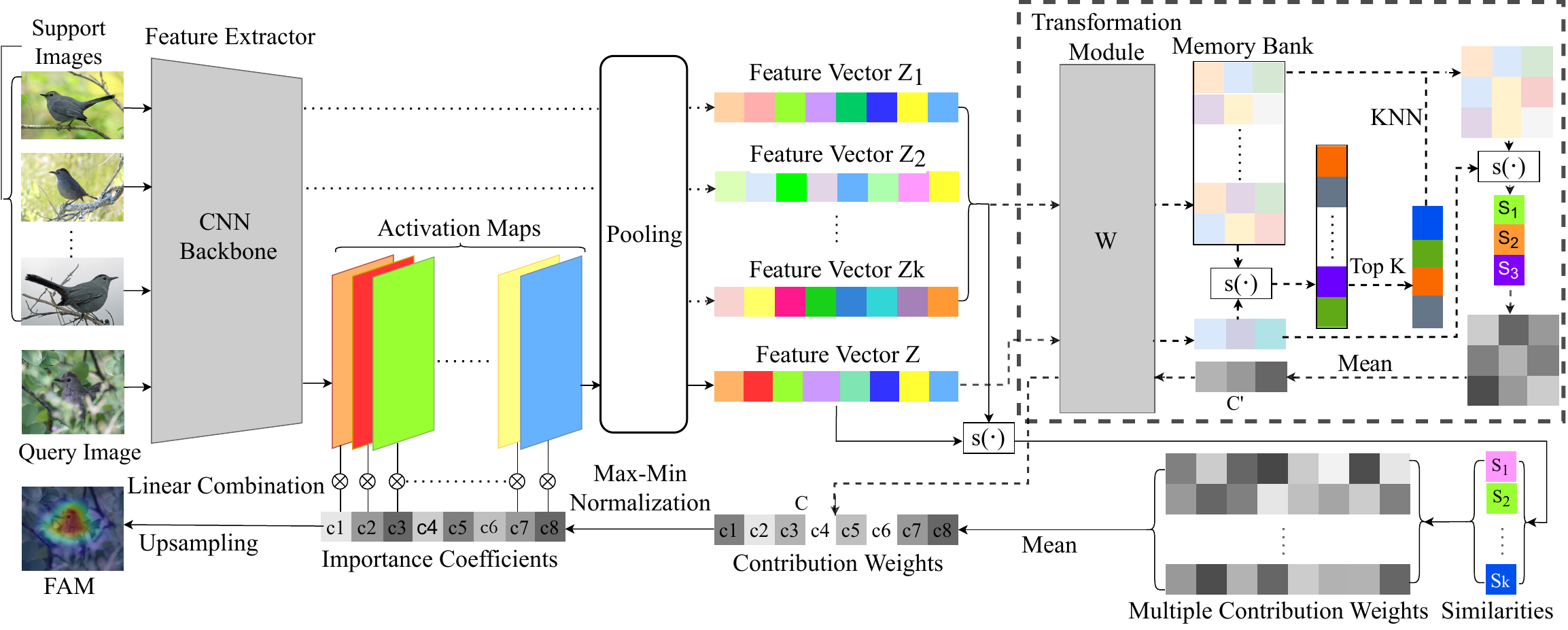}
\caption{The framework of the proposed Feature Activation Map. The dashed box is an optional framework for models with a transformation module and memory bank as a classifier.}
\label{fig1}
\end{figure*}
\section{The Proposed Method}
The importance coefficients are crucial for visualizing explanation maps for a testing sample. All the existing CAM-based methods show that the importance coefficients cannot be obtained without FC layers. They fail to function for visualizing similarity comparison-based deep learning models that do not rely on FC layer as a classifier. To solve this problem, the importance coefficient for each channel is derived from the similarity between samples in the proposed FAM algorithm. 
\subsection{Motivation}
\label{sub4.1}
In convolution neural networks, it is known that the individual convolutional kernel is responsible for generating the activation map corresponding to the channel, so the number of convolution kernels is equal to the number of output channels. Prior studies \cite{ref15LIFTCAM, ref29SHAPValue,ref30LIME,ref31LIFT,ref32} indicate that each activation map can be viewed as an individual semantic feature. In other words, an individual convolution kernel can independently capture a particular semantic feature. The explanation map \cite{ref6CAM,ref7GradCAM,ref12ScoreCAM,ref13RelevanceCAM,ref15LIFTCAM} is essentially a linear combination between activation maps and the corresponding importance coefficients. The importance coefficient determines the influence of the corresponding feature. According to \cite{ref33Theory}, a prediction can be explained by assigning to each feature a number that denotes its influence and the key to explanation is the contribution of individual feature. For the similarity comparison between two images, two activation maps from the same channel are generated by the same convolution kernel. Because a single convolution kernel is responsible for capturing an individual feature, the contribution of a certain channel can reflect the contribution of a feature. Hence, the channel-wise contribution weights play the role of importance coefficients.

\subsection{The Proposed Framework of Feature Activation Map}
\label{fam_framework}
The framework for the proposed feature activation map (FAM) is illustrated in Fig.~\ref{fig1}. Because both semantic and spatial information can be preserved in deeper convolutional layer~\cite{ref8GradCAMplusplus}, the feature map out of the last convolutional layer is employed in the proposed FAM algorithm. The feature map consists of multiple activation maps. The number of activation maps is equal to the number of channels. Therefore, let $f$ denote a CNN backbone as the feature extractor. A testing input image $X$ is fed into backbone network $f$ to obtain feature map $f(X)$$\in$$R^{N\times h\times w}$, where $N$, $h$, and $w$ denote the number of channels, the height and the width of feature map. Thus, the feature map $f(X)$ can be expressed as $f(X)=[A^1,A^2,...,A^N]$, where the $n$-th activation map is denoted by $A^n\in R^{h\times w}$, integer $n=1,2,...,N$. Let $p$ denotes a pooling function that embeds the feature map $f(X)$ into a feature vector $Z=[z_1,z_2,...,z_N]$. The pooling function $p$ can be GAP function, global max pooling (GMP) function~\cite{ref34GMP} and log-sum-exp-pooling function~\cite{ref35LogSum}. Therefore, $Z=p(f(X))$, which is also expressed as $[z_1,z_2,...,z_N]=[p(A^1),p(A^2),...,p(A^N)]$, where $z_n=p(A^n), n=1,2,...,N.$
Let $s$ denotes a similarity metric function, a pair of images $X^A$ and $X^B$ are embedded into $Z^A$ and $Z^B$, the similarity score $S$ is expressed by $S=s(Z^A,Z^B)$, where $s$ can be cosine similarity, Euclidean distance, or other similarity metric functions. The similarity score $S$ is used for prediction. Generally, to reduce the bias from a single sample, the similarity score for decision-making is average value over multiple similarities between a testing sample and other samples from the same category in similarity comparison-based classification paradigm. Thus, assume that there are $K$ images $\{X_{k}^{M}\}_{k=1}^{K}$ from the same category $M$, feature vector $Z_{k}^{M}=p(f(X_{k}^{M}))$ from image $X_{k}^{M}$, feature vector $Z=p(f(X))$ from a testing image $X$, the similarity $S$ for decision-making is obtained by 
\begin{equation}\begin{aligned}
\label{eq11}
S=\frac{1}{K}\sum_{k=1}^{K}s(Z,Z_{k}^{M}).
\end{aligned}\end{equation}

In this subsection, the objective is to decide the channel-wise importance coefficients by using contribution weights from similarity score $S$. The details about calculating contribution weights will be described in the following subsections. Let $C=[c_1,c_2,...,c_N]$, $C\in R^N$ denote the contribution weights from the first channel to the last channel. $c_n$ represents the contribution weight of the $n$-th channel. Contribution weights should be implicitly normalized, which makes comparison and interpretation easier~\cite{ref33Theory}. To reflect the influence of an individual feature, the normalized contribution weight is considered as the importance coefficient $\bar{c}_n$. Thus, for $\forall c_{n}\in C$, we perform max-min normalization to obtain $\bar{c}_n$ by
\begin{equation}\begin{aligned}
\label{eq12}
\bar{c}_n=\frac{c_n-\text{min}(C)}{\text{max}(C)-\text{min}(C)}.
\end{aligned}\end{equation}
Finally, feature activation map $L_{FAM}$ is formed by linearly combining the activation map $A^n$ with its corresponding importance coefficient $\bar{c}_n$ as
\begin{equation}\begin{aligned}
\label{eq13}
L_{FAM}=\sum_{n\text{=}1}^{N}\bar{c}_{n}A^n.
\end{aligned}\end{equation}
From~(\ref{eq13}), the visualization of FAM is generated by using bilinear interpolation as an up-sampling technique.

\subsection{Contribution Weights for Cosine Similarity}
\label{sub4.3}
Any similarity metric can be applied in the proposed framework of FAM. Different similarity metrics cause different ways to calculate channel-wise contribution weights. In this subsection, we provide formulations of computing contribution weights by taking the popular cosine similarity as an example. Let $s$ denote cosine similarity function, for a pair of feature vectors $Z^A$ and $Z^B$, the cosine similarity is calculated as
\begin{equation}\begin{aligned}
\label{eq14}
&s(Z^A,Z^B)=\sum_{n=1}^{N}\frac{z_{n}^Az_{n}^B}{\vert\vert Z^{A}\vert\vert\cdot\vert\vert Z^{B}\vert\vert}, s(Z^A,Z^B)\in[-1,1],\\
\end{aligned}\end{equation}
where $\vert\vert\cdot\vert\vert$ denote $L^2$-norm. Thus, from~(\ref{eq11})~and~(\ref{eq14}), the contribution weight of the $n$-th channel is calculated as
\begin{equation}\begin{aligned}
\label{eq15}
&c_n=\frac{1}{K}\sum_{k=1}^{K}{\frac{z_{n}\cdot z_{n}^{k,M}}{\vert\vert s(Z,Z_{k}^{M})\vert\vert\cdot\vert\vert Z\vert\vert\cdot\vert\vert Z_{k}^{M}\vert\vert}},~c_n\in(-1,1).
\end{aligned}\end{equation}
\subsection{Contribution Weights Transformation}
\label{sub4.4}
Many deep learning methods~ \cite{ref20NPID,ref24MSLoss} change length of the feature vector from last convolutional layer by using FC layers as a transformation module. It is worth noting that the FC layers as a transformation module do not play the role of a classifier, they do not make the decision. Therefore, the feature vector for decision-making is different from the feature vector from the last convolutional layer. The contribution weights, which are obtained from the feature vectors for decision-making, should be inversely transformed to obtain the contribution weights that correspond to the feature vector from the convolutional layer. Let $Z^{\prime}$ denote the feature vector for decision-making, the length of which is marked as $J$, the weights of the transformation module is denoted as $W$$\in$$R^{N\times J}$. It is known that FC layers implement matrix multiplication denoted by ${\times}$, so $Z^{\prime}=Z \times W$, $N$ and $Z$ are defined in Subsection~\ref{fam_framework}. Given the contribution weights $C^{\prime}=[c_1^{\prime},c_2^{\prime},...,c_{J}^{\prime}]$, the contribution $E$ to the similarity score $S$ from a single vector $Z^{\prime}$ can be calculated by
\begin{equation}\begin{aligned}
\label{eq16}
&E=Z^{\prime} \times {C^{\prime}}^{T}=(Z \times W)\times {C^{\prime}}^{T}=Z \times (W\times {C^{\prime}}^{T}). 
\end{aligned}\end{equation}
The contribution to the similarity score $S$ from $Z$ should be the same as $Z^{\prime}$ because both vectors are from the same sample, so the relationship can be expressed by 
\begin{equation}\begin{aligned}
\label{eq17}
&E=Z \times C=Z \times (W \times {C^{\prime}}^{T}),
\end{aligned}\end{equation}
where $C$ is the contribution weights defined in Subsection~\ref{fam_framework}. From~(\ref{eq16}) and~(\ref{eq17}), the contribution weights $C$ is obtained by 
\begin{equation}\begin{aligned}
\label{eq18}
&C=W \times {C^{\prime}}^{T}. 
\end{aligned}\end{equation}

\section{Experiments}
\subsection{Datasets}
The fined-grained image dataset CUB-200-2011 \cite{ref36CUB200} includes 11,788 images from 200 classes. For few-shot image classification tasks, the split strategy is the same as in \cite{ref37SplitCUB200}. The 200 classes are divided into 100, 50, and 50 for training, validation, and testing, respectively. For image retrieval tasks, following the same split strategy as in \cite{ref24MSLoss}, the first 100 classes are used for training and the remaining 100 classes with 5924 images are used as the testing set. The evaluations of the proposed FAM algorithm for both tasks are performed on the testing sets. It is worth noting that the testing class domain is disjoint with the training set, that is, the testing images are from the classes that are not seen before by the models. ImageNet (ILSVRC 2012) \cite{ref38ImageNet2012} consists of 1.28 million training images and 50,000 validation images from 1,000 categories. The evaluation of the proposed FAM algorithm for contrastive learning image classification is performed on the validation set. These datasets provide bounding box annotations which are used for evaluating localization capacity. All experiments are implemented by using Pytorch library with Python 3.8 on NVIDIA RTX 3090 GPU.
\begin{figure*}[!t]
\centering
\setlength{\abovecaptionskip}{5pt}
\setlength{\belowcaptionskip}{5pt}
\includegraphics[width=2.0\columnwidth]{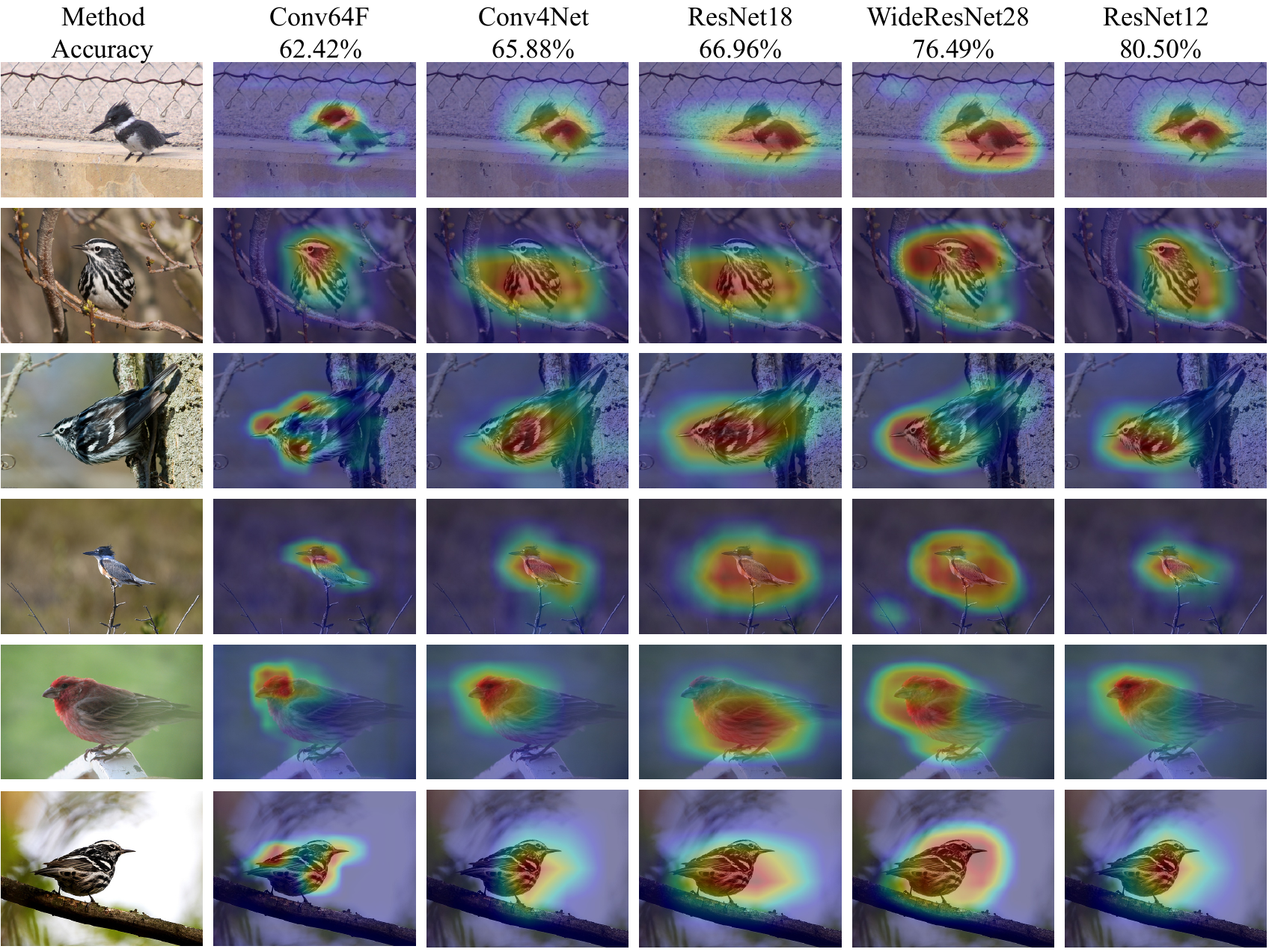}
\caption{Visual explanation maps of the proposed FAM algorithm for five few-shot learning models Conv64F, Conv4Net, ResNet18, WideResNet28, and ResNet12 on CUB-200-2011 dataset (5-way 1-shot).}
\label{fig2}
\end{figure*}

\subsection{Evaluation on Visualizing Few-shot Image Classification CNN Backbones}
{\it{a) Models and Implementation Details:}}~Five most widely used deep learning backbone models in few-shot image classification, Conv4Net \cite{ref22PNet}, Conv64F \cite{ref39Conv64F}, ResNet12 \cite{ref40ResNet12}, ResNet18 \cite{ref2ResNet}, and WideResNet28 \cite{ref41WRN28} are selected as CNN models for evaluating the effectiveness of the proposed FAM algorithm. Global Average Pooling (GAP) is used to embed the feature maps into feature vectors. All models for few-shot image classification use LeakyReLU (slope=0.1) \cite{ref42LeakyReLU} as the activation function. We train the five CNN models using the same episodic training strategy as \cite{ref26CAN}, including data pre-processing strategy, learning rate adjustment rule and optimizer choice. It is worth noting that we use a multi-task learning scheme \cite{ref43MultiTasklearning} to make models focus on the target objects. For validation and testing sets, all images are resized into 84×84 pixels and all pixels are scaled into the range [0,1], and then they are normalized by using mean [0.485, 0.456, 0.406] and standard deviation [0.229, 0.224, 0225]. No further data augmentation technique is applied. Following 5-way K-shot setting, the episode is formed with 5 classes and each class includes K support samples, and 6 and 15 query images are randomly selected for training and inference respectively. During inference, 2,000 episodes are randomly sampled, we apply the proposed FAM algorithm to the five trained CNN models that perform few-shot image classification on the testing set, K is set to 1. The quantitative analyses and qualitative visualizations are performed on the correct predictions. We report the average values of classification accuracy and 95$\%$ confidence interval over the 2000 episodes.

{\it{b) Quantitative and Qualitative Evaluations:}}~
Quantitative analysis includes localization capacity evaluation and faithfulness evaluation. The localization capacity shows how precisely the explanation map can find the discriminative regions on input images. The metrics to measure localization capacity include energy-based point game proportion \cite{ref12ScoreCAM,ref15LIFTCAM} and intersection over union (IoU) \cite{ref6CAM, ref7GradCAM,ref9LayerCAM}. Energy-based point game proportion shows how much energy of the explanation map falls into the bounding box of the target object. It can be formulated as follow,
\begin{equation}\begin{aligned}
\label{eq19Proportion}
\text{Proportion}=\frac{\sum_{(i,j)\in \text{bbox}}(\text{norm}(\text{up}(L_{FAM})))_{(i,j)}}{\sum_{(i,j)\in\land}(\text{norm}(\text{up}(L_{FAM})))_{(i,j)}},
\end{aligned}\end{equation}
where $\land$ denotes the size of the original image $X$, bbox represents the ground truth bounding box, and $(i,j)$ denotes the location of a pixel. Intersection over union (IoU) involves the estimated bounding box. To generate an estimated bounding box from the proposed FAM algorithm, following \cite{ref6CAM, ref7GradCAM,ref9LayerCAM}, a simple threshold technique is used to segment the saliency map. We first binarize the saliency map with the threshold of the max value of saliency map. Second a bounding box is drawn, which covers the largest connected segments of pixels. IoU is calculated by the following, 
\begin{equation}\begin{aligned}
\label{eq20IoU}
\text{IoU}=\frac{\sum_{ (i,j)\in {(\text{bbox}_e \bigcap \text{bbox}_g})}{1}}{\sum_{(i,j)\in{(\text{bbox}_e \bigcup \text{bbox}_g})}{1}},
\end{aligned}\end{equation}
where $\text{bbox}_e$ denotes the estimated bounding box while $\text{bbox}_g$ denotes the ground truth bounding box. Following \cite{ref6CAM}, the threshold is set to 0.2 in our experiments. Both metrics are adopted to assess the localization capacity of the proposed FAM algorithm. The annotated bounding boxes provided by CUB-200-2011 dataset are used as ground-truth labels. During inference, we compute IoU and proportion for a query image that is correctly classified. For every episode, we calculate the average IoU and proportion of all correct classifications as well as accuracy per episode. The means of 2,000 episodes of average IoU and proportion are listed in Table~\ref{t1}. The results in Table~\ref{t1}  show that ResNet-based models have better localization capacity than Conv64F or Conv4Net.
\begin{table}[!t]
\caption{The evaluation for localization capacity. Higher value is better for IoU and Energy-based point game proportion}
\renewcommand{\arraystretch}{1.2}
\centering
{{\begin{tabular}{ccc}
 \hline
      {Models}  & {Proportion } & {IoU}\\
 \hline
 Conv64F           &36.00$\%$ &    45.02$\%$\\
 Conv4Net          &38.48$\%$ &    43.45$\%$\\
 ResNet18          &44.75$\%$ &    47.35$\%$\\
 WideResNet28      &43.38$\%$ &    49.72$\%$\\
 ResNet12          &40.96$\%$ &    48.63$\%$\\
 \hline
\end{tabular}
}}
\label{t1}
\end{table}
\begin{table}[!t]
\caption{The evaluation on faithfulness. Lower is better for Average Drop. Higher is better for Increase in Confidence}
\renewcommand{\arraystretch}{1.2}
\centering
{{\begin{tabular}{ccc}
 \hline
      {Models}  & {Average Drop  } & {Increase in Confidence}\\
 \hline
 Conv64F           &14.67$\%$ &    39.79$\%$\\
 Conv4Net          &14.36$\%$ &    44.59$\%$\\
 ResNet18          &13.98$\%$ &    45.13$\%$\\
 WideResNet28      & 7.97$\%$ &    41.28$\%$\\
 ResNet12          &11.68$\%$ &    43.26$\%$\\
 \hline
\end{tabular}
}}
\label{t2}
\end{table}
\begin{figure*}[!t]
\centering
\setlength{\abovecaptionskip}{5pt}
\setlength{\belowcaptionskip}{5pt}
\includegraphics[width=2.0\columnwidth]{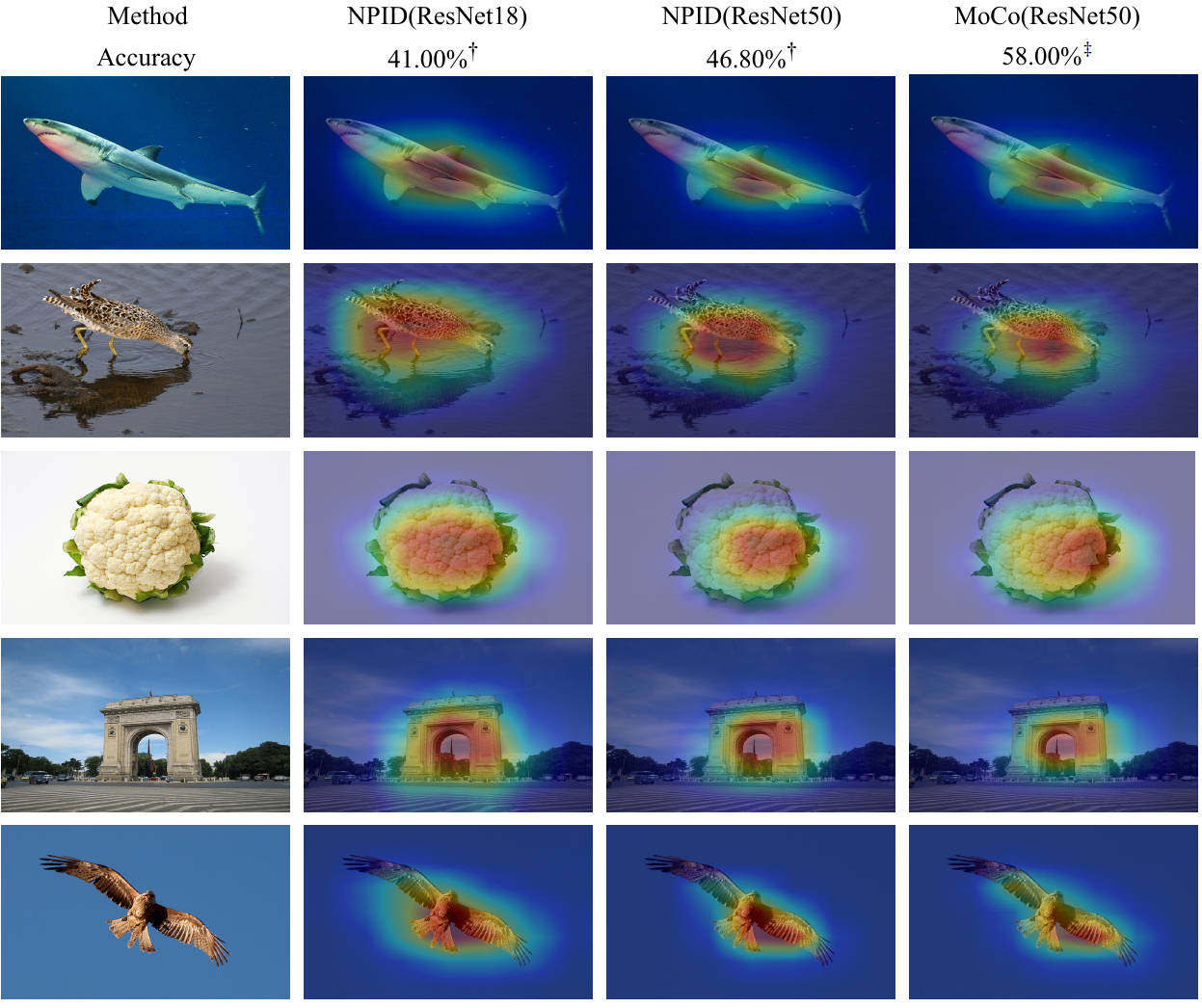}
\caption{Visual explanation maps of the proposed FAM algorithm for NPID (ResNet18), NPID (ResNet50), and MoCo (ResNet50) on ImageNet2012 dataset.~\dag~and~\ddag~denote that the results are from~\cite{ref20NPID} and~\cite{ref23MoCo} respectively.}
\label{fig3}
\end{figure*}

The faithfulness measures how important the regions that the explanation map highlights will be. Therefore, the faithfulness evaluation has been widely performed for interpretable visualization methods \cite{ref7GradCAM,ref8GradCAMplusplus,ref12ScoreCAM,ref13RelevanceCAM,ref15LIFTCAM}. Following \cite{ref8GradCAMplusplus, ref12ScoreCAM,ref13RelevanceCAM,ref15LIFTCAM}, average drop (AD) and increase in confidence (IC) are chosen as the faithfulness metrics to evaluate the faithfulness of the proposed FAM method. Both metrics are defined as the following, 
\begin{equation}\begin{aligned}
\label{eq21AD}
\text{AD}=\frac{1}{N}\sum_{i=1}^{N}\frac{\text{max}(0,s_i-s'_{i})}{s_i}\times100,
\end{aligned}\end{equation}
and
\begin{equation}\begin{aligned}
\label{eq22IC}
\text{IC}=\frac{1}{N}\sum_{i=1}^{N}1_{(s_i<s'_{i})}\times100,
\end{aligned}\end{equation}
where $s_i$ denotes the similarity score between the $i$-th query image and the corresponding support image and $s'_i$ denote the similarity score between the $i$-the query image masked by the saliency map and the same support image. $N$ indicates the number of query images per episode. We calculate the means of AD and IC for 2,000 episodes, which are listed in Table~\ref{t2}. Table~\ref{t2}  shows that ResNet-based models have a better performance in most cases except the IC on Conv4Net, where Conv4Net outperforms ResNet12 and WideResNet28. For qualitative visualization, we randomly select four episodes from the testing set by using the python function "enumerate (dataloader)". The proposed FAM algorithm is used to generate the explanation maps on the images, for which the five few-shot models consistently make the correct classification. Fig.~\ref{fig2} shows examples of FAM visualization (more samples are shown in the appendix). The head of Fig.~\ref{fig2} also displays the classification accuracies. It can be observed that with the improvement of the classification accuracy column by column from left to right, the regions highlighted by FAM look more focused on targeted objects in accordance with human observation.
\begin{table}[!t]
\caption{The evaluation for localization capacity for NPID (ResNet18)~\cite{ref20NPID}, NPID (ResNet50)~\cite{ref20NPID} and MoCo (ResNet50)~\cite{ref23MoCo}}
\renewcommand{\arraystretch}{1.2}
\centering
{{\begin{tabular}{ccc}
 \hline
      {Models}  & {Proportion} & {IoU}\\
 \hline
 NPID (ResNet18)     &54.48$\%$ &    54.01$\%$\\
 NPID (ResNet50)     &54.29$\%$ &    52.81$\%$\\
 MoCo (ResNet50)     &54.33$\%$ &    52.68$\%$\\
 \hline
\end{tabular}
}}
\label{t3}
\end{table}
\subsection{Evaluation on Visualizing Contrastive Learning Image Classification CNN Models}
{\it{a) Implement Details:}}~The three pretrained contrastive learning  models\footnote{http://github.com/zhirongw/lemniscate.pytorch} (i.e., two non-parametric instance discriminative (NPID) models~\cite{ref20NPID} and a momentum contrast (MoCo) model~\cite{ref23MoCo}) are used to perform image classification on ImageNet validation set. The off-the-shelf memory bank from the corresponding model plays the role of KNN classifier, K is set to 200~\cite{ref20NPID}. Each image is resized into 224$\times$224 pixels and then normalized by using mean [0.485, 0.456, 0.406] and standard deviation [0.229, 0.224, 0225].

{\it{b) Quantitative and Qualitative Evaluations:}}~We test the 50,000 images in the validation set of ImageNet using the three pretrained models. To evaluate localization capacity, we calculate the average IoU and proportion for all correctly classified images with the assistance of the annotated bounding boxes provided by the ImageNet (ILSVRC 2012)~\cite{ref38ImageNet2012}. The localization performance for the three models are reported in Table~\ref{t3}. For qualitative evaluation, we use the three pretrained contrastive learning models to classify all the 50,000 images from the validation set. The FAMs of the correctly classified images by the three models are displayed for our qualitative visual evaluation. Here, we randomly select examples using the command of "random.sample()" and display the images and their FAM saliency visualizations for the three contrastive learning CNN models in Fig.~\ref{fig3} (additional samples are displayed in the appendix).

\begin{figure*}[!t]
\centering
\setlength{\abovecaptionskip}{5pt}
\setlength{\belowcaptionskip}{5pt}
\includegraphics[width=2.0\columnwidth]{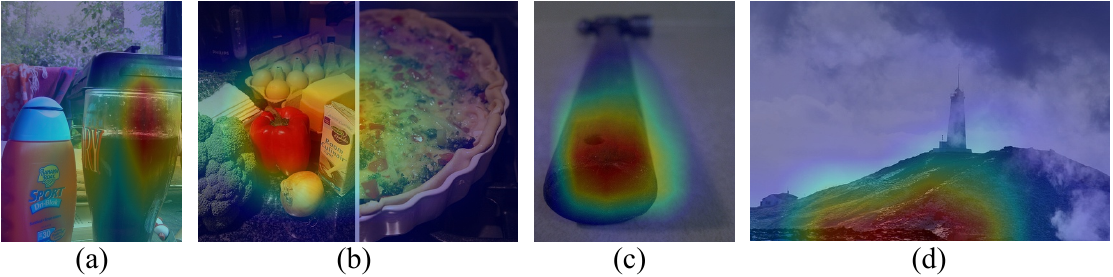}
\caption{Examples of how FAM effectively explains incorrect classifications made by the NPID (ResNet18) model. The NPID (ResNet18) predicts (a) as a beer glass because its attention, as indicated by FAM, focuses on the beer glass rather than the sunscreen (ground truth label); (b) is classified as a bell pepper because its attention is located on the bell pepper rather than the broccoli (ground truth label); (c) is categorized as a matchstick rather than a hammer (ground truth label); (d) is recognized as the drop cliff rather than a lighthouse (ground truth label).}
\label{fig4}
\end{figure*}

To analyse the reason that a model makes a wrong prediction, we display FAMs of testing images that are incorrectly classified by NPID (ResNet18), as shown in Fig.~\ref{fig4}. Because the images in ImageNet dataset may contain multiple objects (e.g., beer glass and sunscreen in Fig.~\ref{fig4}(a)) in a single image, but each image only has one ground truth label (i.e., sunscreen for the image of Fig.~\ref{fig4}(a)), it is interesting to see if the proposed FAM can interpret how a model makes incorrect classification decision. Fig.~\ref{fig4} reveals the predictions are consistent with the region indicated by the FAM explanation maps. For example, Fig.~\ref{fig4}(b) is classified by the CNN model as a "bell pepper" because its attention is located on the bell pepper as shown by the FAM explanation map, rather than the broccoli (the ground truth label of the image). It demonstrates the effectiveness of the proposed FAM as an interpretation tool for understanding and explaining the (incorrect classification) decisions made by the deep learning models.
\begin{table}[!t]
\caption{The evaluation for localization capacity of the proposed FAM algorithm for Margin (Inception)~\cite{ref28MarginLoss} and MS (Inception)~\cite{ref24MSLoss}}
\renewcommand{\arraystretch}{1.2}
\centering
{{\begin{tabular}{ccc}
 \hline
      {Methods}  & {Proportion} & {IoU}\\
 \hline
 Margin (Inception)      &46.27$\%$ &    52.23$\%$\\
 Multi-Similarity (Inception) &46.64$\%$ & 52.29$\%$\\
 \hline
\end{tabular}
}}
\label{t4}
\end{table}
\subsection{Evaluation on Visualizing Image Retrieval CNN Models}
Two CNN models widely used as benchmarks for image retrieval, multi-similarity model (MS model)~\cite{ref24MSLoss} and Margin model~\cite{ref28MarginLoss}, are used to examine the effectiveness of the proposed FAM. We use the publicly released codes\footnote{https://github.com/msight-tech/research-ms-loss} of MS and Margin models~\cite{ref24MSLoss, ref28MarginLoss} in our experiments, which provide us the same Recall@1 results for the MS and the Margin models as given in~\cite{ref24MSLoss}. Recall@1 is the percentage of testing images correctly retrieved in which the most similar image (top 1) retrieved from the database belongs to the same class of the testing image.~We calculate the localization capacity of the proposed FAM algorithm for the MS and Margin models using the images correctly recognized by the model respectively. The results are summarized in Table~\ref{t4}. To qualitatively analyze the visual explanation of FAM, we randomly select examples by using the function of "random.sample()" from the testing images that both models make the correct decision. These images and their explanation maps generated by the proposed FAM algorithm are shown in Fig.~\ref{fig5} (more samples are displayed in the appendix).~The visual results illustrate that the proposed FAM method locates the target birds and highlights the discriminative regions where the models extract feature information for correct image retrieval.~It is interesting to see that the explanation maps show small difference between MS and Margin models on the same testing image. Different from the five CNN models for few-shot learning and three CNN models for contrastive learning that use different backbones, both MS and Margin models use the same backbone. This may be the reason that the difference of FAMs between MS and Margin models in Fig.~\ref{fig5} are smaller than those among the five models in Fig.~\ref{fig2} and those among the three models in Fig.~\ref{fig3}.
\begin{figure*}[!t]
\centering
\setlength{\abovecaptionskip}{5pt}
\setlength{\belowcaptionskip}{5pt}
\includegraphics[width=2.0\columnwidth]{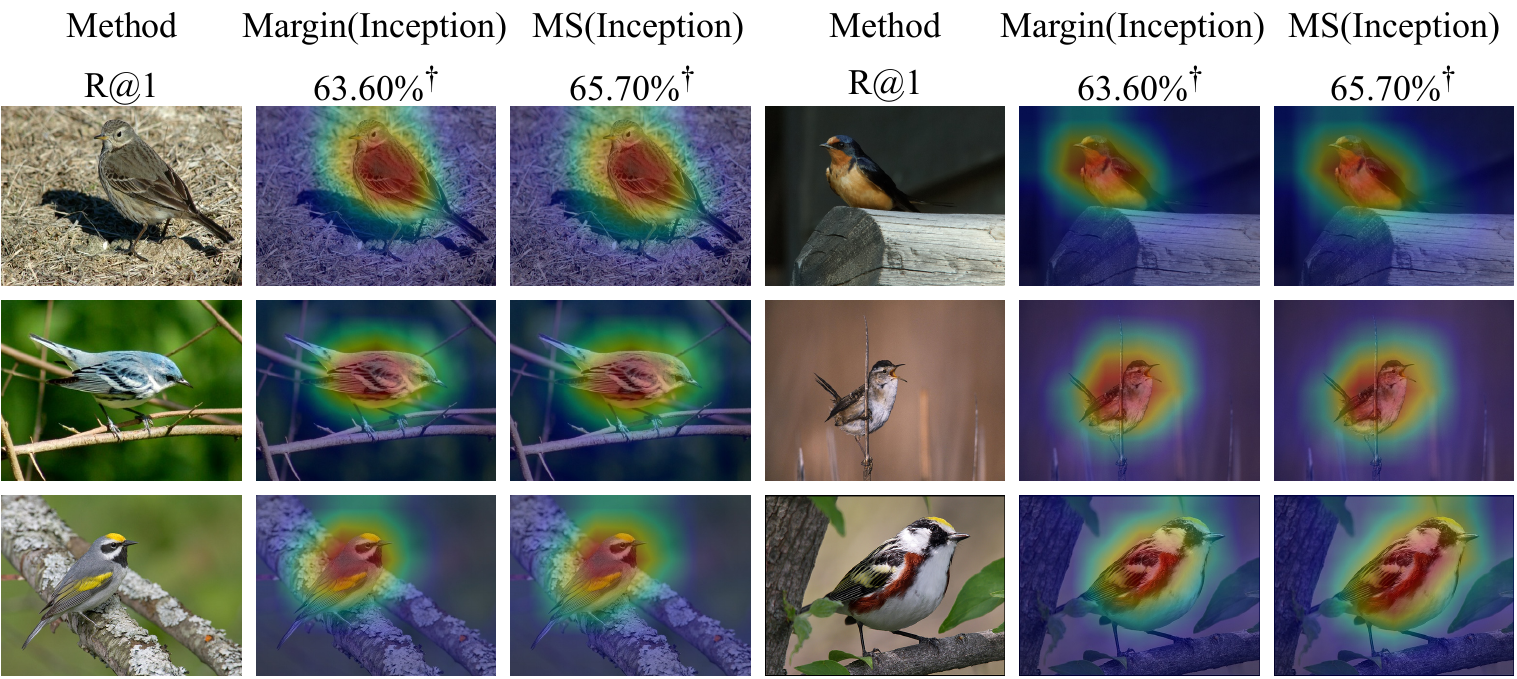}
\caption{The explanation maps generated by the proposed FAM algorithm for Margin and Multi-Similarity models on CUB-200-2011 dataset.~\dag~denotes that the results are from~\cite{ref24MSLoss}.}
\label{fig5}
\end{figure*}

\section{Conclusion}
In this paper, we present a novel model-agnostic visual explanation algorithm named feature activation map (FAM), bridging the gap of visualizing explanation maps for FC-free deep learning models in image classification tasks. The proposed FAM algorithm determines the importance coefficients by using the contribution weights to the similarity score. The importance coefficients are linearly combined with the activation maps from the last convolutional layer, forming the explanation maps for visualization. The qualitative and quantitative experiments conducted on 10 widely used CNN models demonstrate the effectiveness of the proposed FAM for few-shot image classification, contrastive learning image classification and image retrieval tasks. In future, we hope that FAM can be used broadly for explaining deep learning models in large.


{\appendix[]
\section{FAM Visualization}
In the appendix, more explanation maps generated by the proposed FAM for the five few-shot learning backbones are visualized in Fig.~\ref{fig6} and Fig.~\ref{fig7}. More explanation maps generated by the proposed FAM algorithm for the three contrastive learning models are shown in Fig.~\ref{fig8}. Fig.~\ref{fig9} display more explanation maps generated by the proposed FAM algorithm for the two image retrieval CNN models.

\begin{figure*}
\centering
\setlength{\abovecaptionskip}{5pt}
\setlength{\belowcaptionskip}{5pt}
\includegraphics[width=2.0\columnwidth]{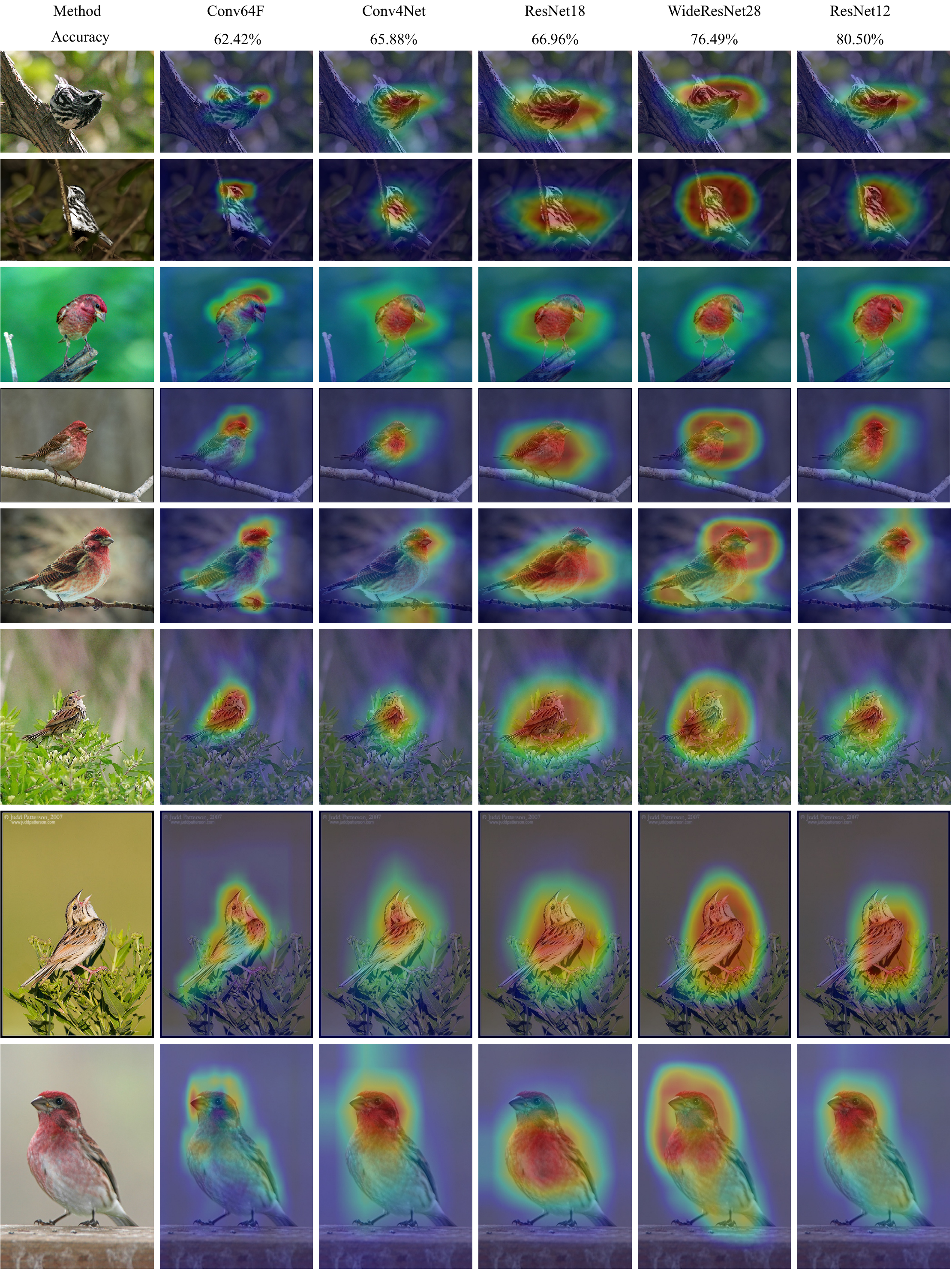}
\caption{More explanation maps generated by the proposed FAM algorithm for Conv64F, Conv4Net, ResNet18, WideResNet28,and ResNet12 on CUB-200-2011 dataset (5-way 1-shot).}
\label{fig6}
\end{figure*}
\begin{figure*}
\centering
\setlength{\abovecaptionskip}{5pt}
\setlength{\belowcaptionskip}{5pt}
\includegraphics[width=2.0\columnwidth]{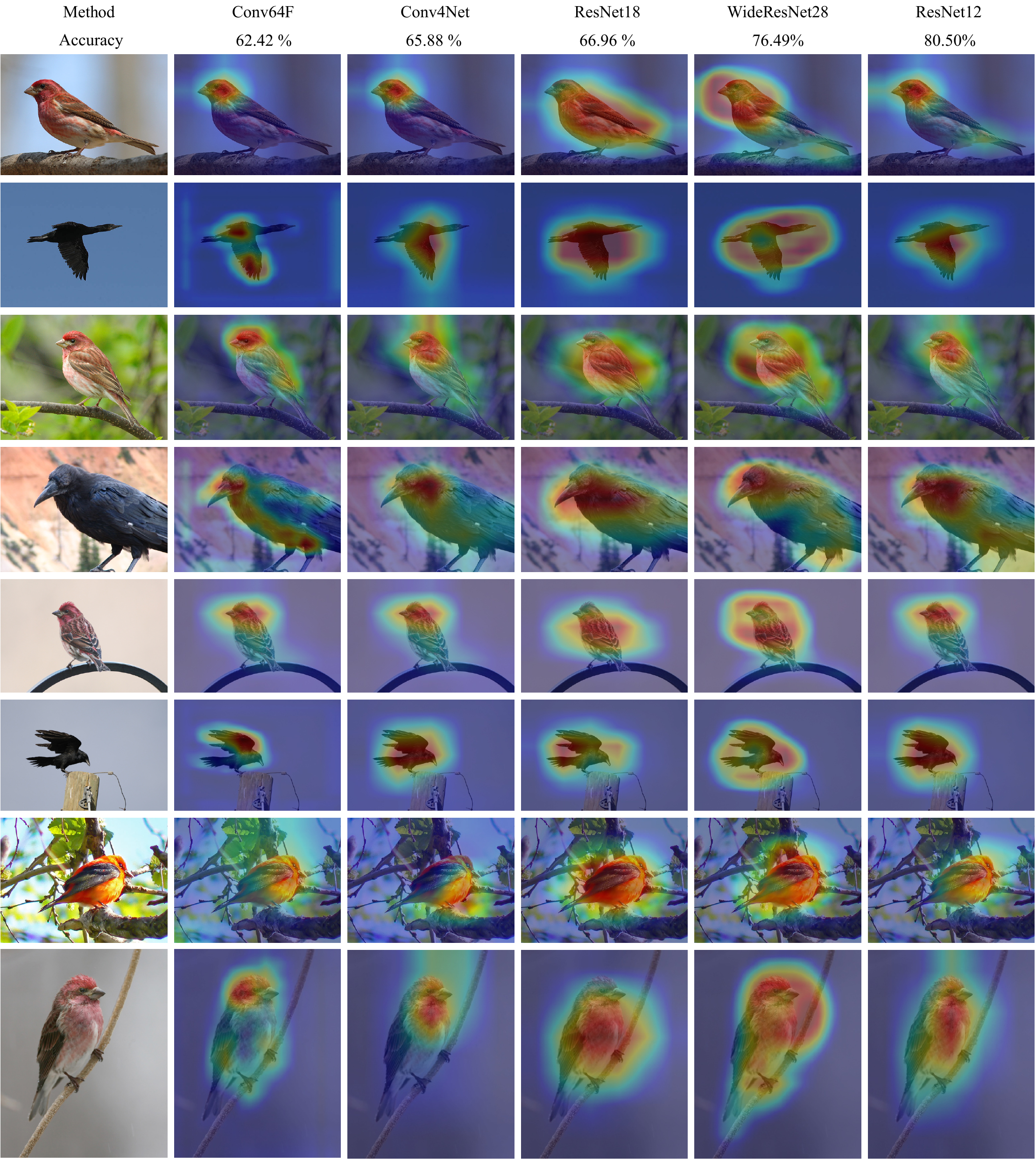}
\caption{More explanation maps generated by the proposed FAM algorithm for Conv64F, Conv4Net, ResNet18, WideResNet28, and ResNet12 on CUB-200-2011 dataset (5-way 1-shot).}
\label{fig7}
\end{figure*}
\begin{figure*}
\centering
\setlength{\abovecaptionskip}{5pt}
\setlength{\belowcaptionskip}{5pt}
\includegraphics[width=2.0\columnwidth]{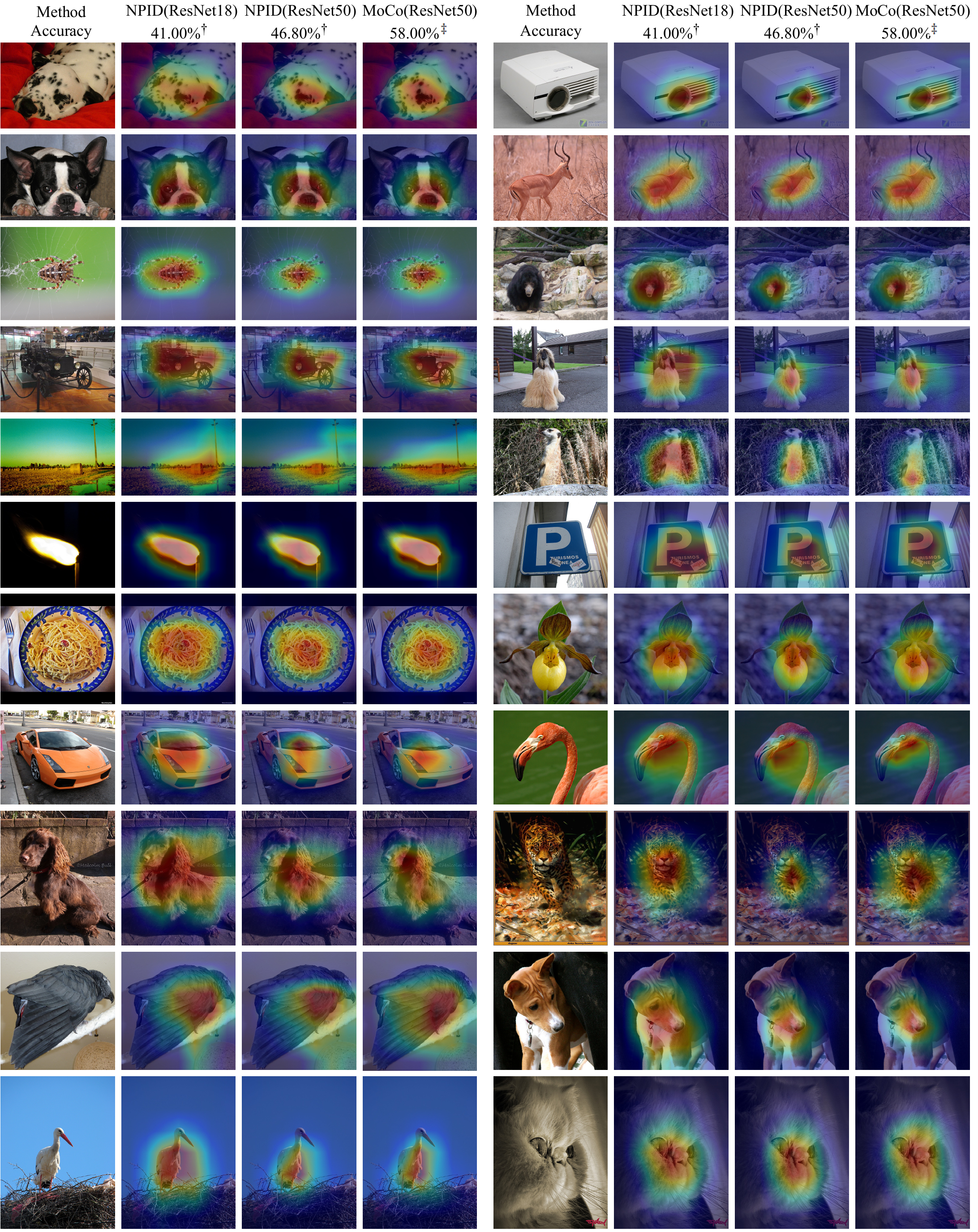}
\caption{More explanation maps generated by the proposed FAM algorithm for NPID (ResNet18), NPID (ResNet50), and MoCo (ResNet50) on ImageNet2012 dataset.~\dag~and~\ddag~denote that the results are from~\cite{ref20NPID} and~\cite{ref23MoCo} respectively.}
\label{fig8}
\end{figure*}

\begin{figure*}
\centering
\setlength{\abovecaptionskip}{5pt}
\setlength{\belowcaptionskip}{5pt}
\includegraphics[width=2.0\columnwidth]{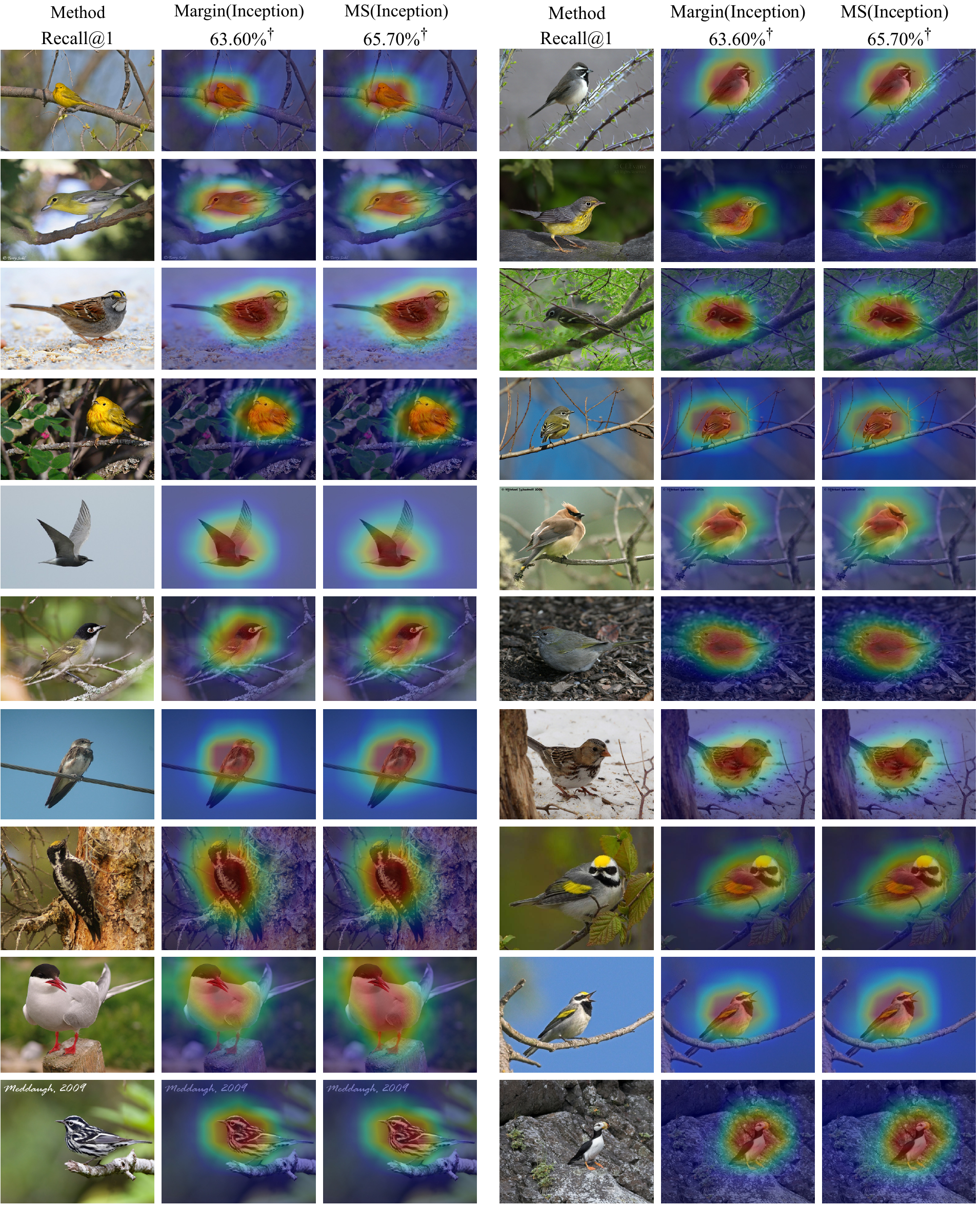}
\caption{More explanation maps generated by the proposed FAM algorithm for Margin and Multi-Similarity models on CUB-200-2011 dataset.~\dag~denotes that the results are from~\cite{ref24MSLoss}.}
\label{fig9}
\end{figure*}

}




\vspace{11pt}


\begin{IEEEbiography}[{\includegraphics[width=1in,height=1.25in,clip,keepaspectratio]{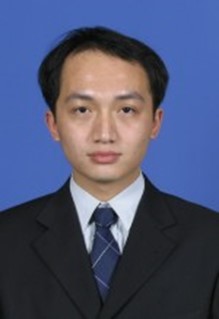}}]{Yi Liao} received the B.S. degree in clinical medicine from Fudan University, Shanghai, China, in 2007, and M.S. degree of computer science from Queensland University of Technology, Brisbane, Australia, in 2019. He is currently pursuing the Ph.D. degree of Artificial Intelligence at School of Engineering and Built Environment, Griffith University, Brisbane, Australia. 

His research interests include deep learning, image classification and object recognition.
\end{IEEEbiography}
\vspace{11pt}
\begin{IEEEbiography}
[{\includegraphics[width=1in,height=1.25in,clip,keepaspectratio]{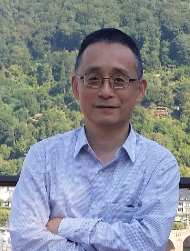}}]{Yongsheng Gao} received the B.Sc. and M.Sc. degrees in electronic engineering from Zhejiang University, Hangzhou, China, in 1985 and 1988, respectively, and the Ph.D. degree in computer engineering from Nanyang Technological University, Singapore. He is currently a Professor with the School of Engineering and Built Environment, Griffith University, and the Director of ARC Research Hub for Driving Farming Productivity and Disease Prevention, Australia. He had been the Leader of Biosecurity Group, Queensland Research Laboratory, National ICT Australia (ARC Centre of Excellence), a consultant of Panasonic Singapore Laboratories, and an Assistant Professor in the School of Computer Engineering, Nanyang Technological University, Singapore. 

His research interests include smart farming, machine vision for agriculture, biosecurity, face recognition, biometrics, image retrieval, computer vision, pattern recognition, environmental informatics, and medical imaging. 
\end{IEEEbiography}

\vspace{11pt}
\begin{IEEEbiography}
[{\includegraphics[width=1in,height=1.25in,clip,keepaspectratio]{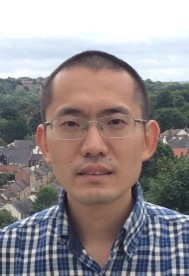}}]{Weichuan Zhang} received the MS degree in signal and information processing from the Southwest Jiaotong University in China and the PhD degree in signal and information processing in National Lab of Radar Signal Processing, Xidian University, China. He is currently a research fellow at Griffith University, Brisbane, Australia. 

His research interests include computer vision, image analysis, and pattern recognition.
\end{IEEEbiography}

\vspace{11pt}
\vfill
\end{document}